\documentclass[a4paper,conference]{IEEEtran}
\usepackage{cite}

\ifCLASSINFOpdf
   \usepackage[pdftex]{graphicx}
   \graphicspath{{./graphics/}}
   \DeclareGraphicsExtensions{.pdf,.jpeg,.png}
\else
   \usepackage[dvips]{graphicx}
   \graphicspath{{./eps/}}
   \DeclareGraphicsExtensions{.eps}
\fi
\usepackage{amsmath}
\usepackage{array}

\ifCLASSOPTIONcompsoc
  \usepackage[caption=false,font=normalsize,labelfont=sf,textfont=sf]{subfig}
\else
  \usepackage[caption=false,font=footnotesize]{subfig}
\fi
\usepackage{fixltx2e}

\usepackage{stfloats}
\usepackage{url}

\usepackage{booktabs} 
\usepackage{tabularx} 
\usepackage{multirow, makecell} 
\usepackage{caption}
\usepackage{csquotes} 
\usepackage{dsfont}
\usepackage{amssymb}
\usepackage{algorithm} 
\usepackage[noend]{algpseudocode} 

\begin{document}
%
\title{ContraCluster: Learning to Classify without Labels \\
	 by Contrastive Self-Supervision and Prototype-Based Semi-Supervision}

\author{\IEEEauthorblockN{Seongho Joe, Byoungjip Kim, Hoyoung Kang, Kyoungwon Park, Bogun Kim, Jaeseon Park, Joonseok Lee, \\
		Youngjune Gwon}
\IEEEauthorblockA{Samsung SDS, Seoul, South Korea \\
	Email: \{drizzle.cho, bjip.kim, hoyoung.kang, kw621.park, bogun0.kim, jaeseon.park, js1985.lee, \\
	gyj.gwon\}@samsung.com}}


%


\maketitle

\begin{abstract}
The recent advances in representation learning inspire us to take on the challenging problem of unsupervised image classification tasks in a principled way. We propose ContraCluster, an unsupervised image classification method that combines clustering with the power of contrastive self-supervised learning. ContraCluster consists of three stages: (1) contrastive self-supervised pre-training (CPT), (2) contrastive prototype sampling (CPS), and (3) prototype-based semi-supervised fine-tuning (PB-SFT). CPS can select highly accurate, categorically prototypical images in an embedding space learned by contrastive learning. We use sampled prototypes as noisy labeled data to perform semi-supervised fine-tuning (PB-SFT), leveraging small prototypes and large unlabeled data to further enhance the accuracy. We demonstrate empirically that ContraCluster achieves new state-of-the-art results for standard benchmark datasets including CIFAR-10, STL-10, and ImageNet-10. For example, ContraCluster achieves about 90.8\% accuracy for CIFAR-10, which outperforms DAC (52.2\%), IIC (61.7\%), and SCAN (87.6\%) by a large margin. Without any labels, ContraCluster can achieve a 90.8\% accuracy that is comparable to 95.8\% by the best supervised counterpart.
\end{abstract}


 \ifCLASSOPTIONpeerreview
 \begin{center} \bfseries EDICS Category: 3-BBND \end{center}
 \fi
%
\IEEEpeerreviewmaketitle

\section{Introduction}

Supervised learning approaches in deep learning have shown to provide a human-level performance in computer vision tasks such as image classification \cite{he2016deep} and object detection \cite{mask2017kaiming}. Unsupervised learning, in contrast, has long been considered too challenging for discriminative machine learning tasks \cite{chang2017deep, ji2019invariant} and difficult to provide an accuracy comparable to that of supervised learning.
\begin{figure}[t]
    \begin{center}
    \includegraphics[width=0.7\linewidth]{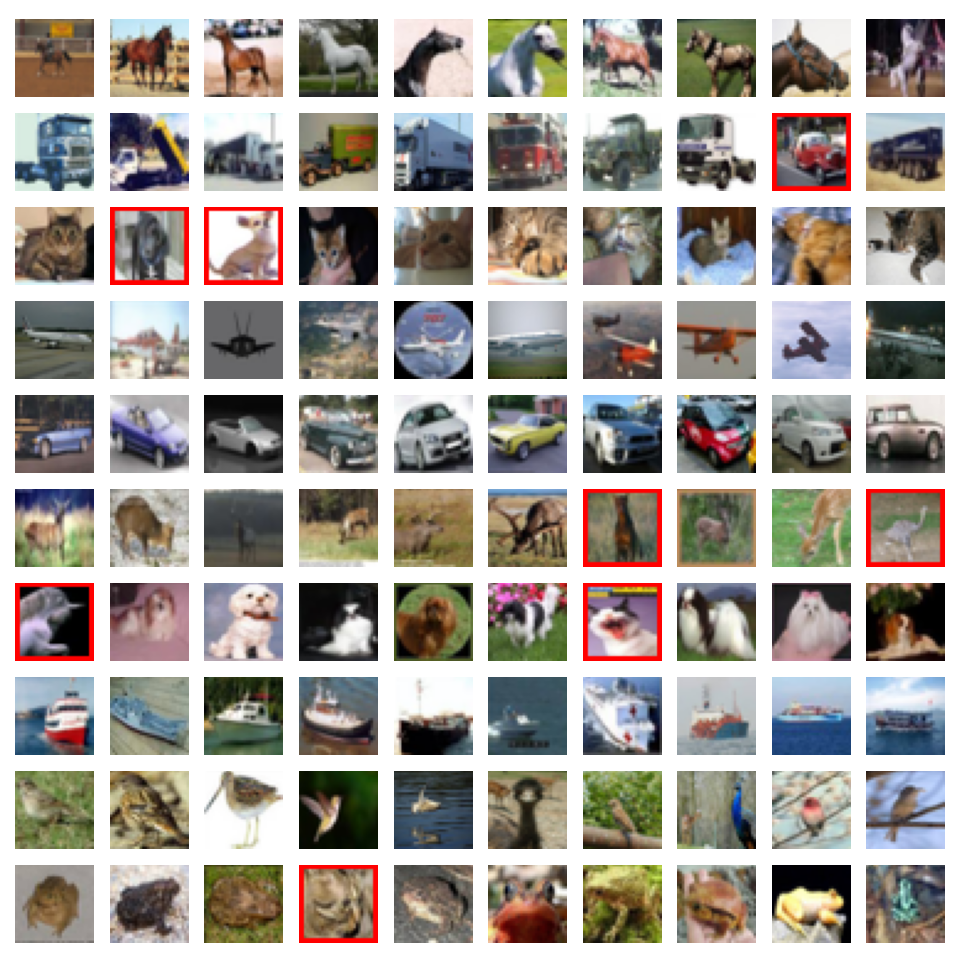}
    \end{center}
    \caption{An example clustering result of ContraCluster for CIFAR-10. For visualization, ten images are randomly sampled from the final clustering result. Each rows is a cluster discovered by ContraCluster. Red rectangles denotes misclassified images in each cluster. In this example, there are only 8 errors over 100 samples. This example approximately shows that ContraCluster provides such high accuracy (i.e., 90.8\%) without labels.}
    \label{fig:contracluster_main_result}
\end{figure} 

\begin{figure}[t]
    \begin{center}
    \includegraphics[width=0.95\linewidth]{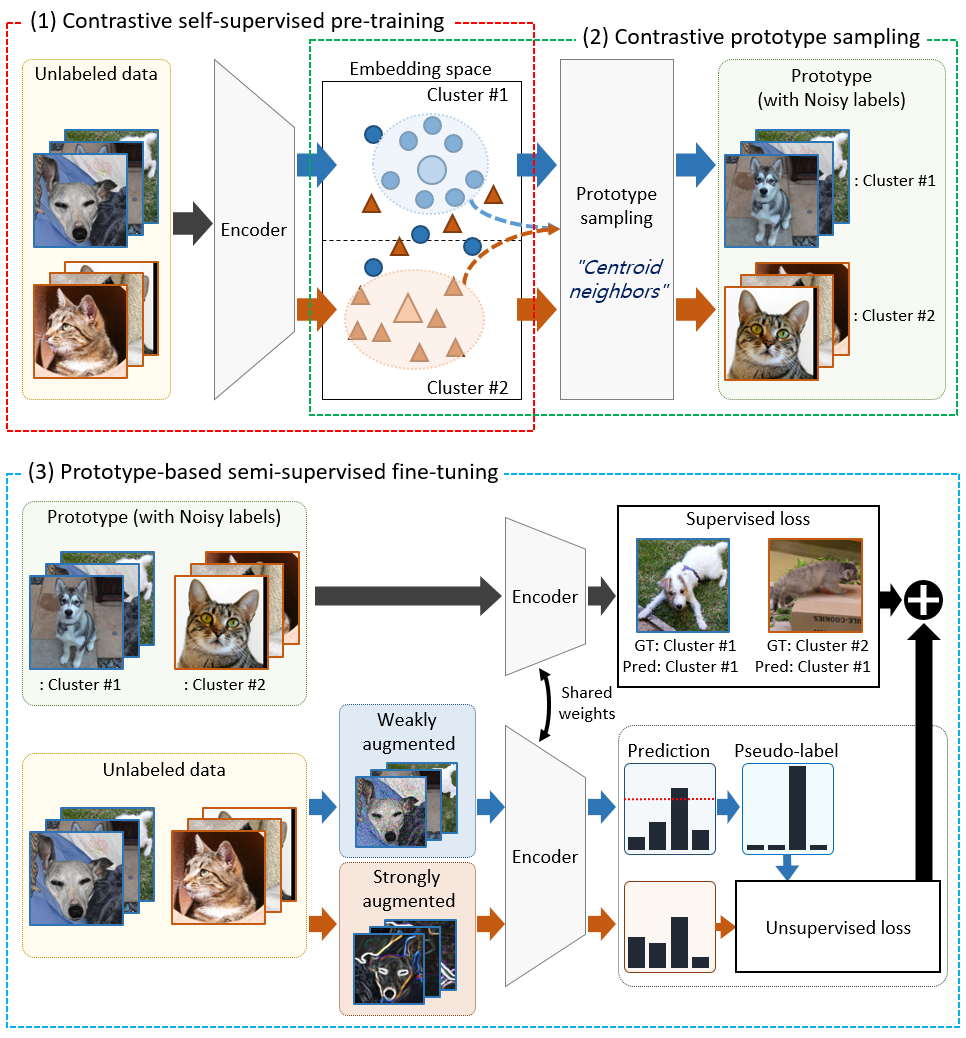}
    \end{center}
    \caption{The overview of ContraCluster. It consists of three stages: (1) contrastive self-supervised pre-training (CPT), (2) contrastive prototype sampling (CPS), and (3) prototype-based semi-supervised fine-tuning (PB-SFT). It selects highly accurate prototypical samples (i.e., prototypes) from an embedding space learned by contrastive self-supervised pre-training. They are used as noisy labeled data in PB-SFT. Note that, in the proposed pipeline, ContraCluster does not use any human-labeled data to classify images.}
    \label{fig:contracluster_overview}
\end{figure} 

But advances in self-supervised representation learning makes it possible for pre-train models to learn general features from unlabeled data by generating pretext tasks \cite{doersch2015unsupervised, noroozi2016unsupervised, zhang2016colorful, gidaris2018unsupervised}. Recently, contrastive self-supervised representation learning such as CPC \cite{oord2018representation}, DIM \cite{hjelm2019learning}, MoCo \cite{he2020momentum}, and SimCLR \cite{chen2020simple} has significantly enhanced the quality of learned representations by using the InfoMax principle \cite{tschannen2020mutual}. 

It enables the paradigm of unsupervised pre-training followed by fine-tuning with few labels \cite{chen2020simple}. For example, SimCLRv2 \cite{chen2020big} propose a distillation stage that takes place after supervised fine-tuning. S4L \cite{zhai2019s4l} and SelfMatch \cite{kim2020selfmatch} show that semi-supervised fine-tuning further enhance the accuracy and label efficiency.

Combining them, end-to-end unsupervised learning schemes emerge, such as SCAN \cite{van2020scan} and RUC \cite{park2020improving}. SCAN consists of three stages: (1) contrastive self-supervised pre-training (SimCLR), (2) fine-tuning with SCAN loss, and (3) fine-tuning with self-labeling. The main idea of SCAN is the SCAN loss neighborhood consistency that encourages the model to make consistent predictions between a sample and its neighboring samples. RUC is based on SCAN, and proposes additional fine-tuning stages: (4) clean sample selection based on confidence scores, and (5) semi-supervised fine-tuning with MixMatch \cite{berthelot2019mixmatch}, which is known as interpolation consistency regularization.

They are robust methods and show high performances. Some weak points, however, exist. the SCAN loss has high computational complexity, as it needs to select k neighboring samples for each sample at every optimizing steps. And MixMatch used in RUC suffers from relatively low accuracy when only small number of labeled data available. Moreover, SCAN uses k-Nearest Neighbor (k-NN) for sematic clustering, and RUC utilizes both k-NN and confidence scores. But, k-NN is hard to reveal global structure of embedding space. It could mix noisy cluster boundaries in the pseudo-label, resulting memorization. It is also well known that confidence scores provided by neural networks are not good estimates for the uncertainty of class assignment.

We introduce ContraCluster, an unsupervised image classification method that leverages the advances of contrastive self-supervised learning via clustering. Figure \ref{fig:contracluster_main_result} shows an example clustering result of ContraCluster. As shown in Figure \ref{fig:contracluster_overview}, ContraCluster consists of three stages: (1) contrastive self-supervised pre-training (CPT), (2) contrastive prototype sampling (CPS), and (3) prototype-based semi-supervised fine-tuning (PB-SFT). 

In the first stage, we aim to discover a linearly separable embedding space by using only unlabeled data. To achieve this goal, we perform \emph{contrastive self-supervised pre-training} (CPT). Among many promising methods, we adopt SimCLR \cite{chen2020simple}. Unlike SCAN, we directly use SimCLR results, resulting simpler pipeline. (for details see \ref{sec:stage1}). 

For the second stage, we develop \emph{contrastive prototype sampling} (CPS) that selects prototypical images that are highly categorical from the learned embedding space in the first stage. (see Figure \ref{fig:silhouette_scores_prototype_accuracy}). Conceptually, highly categorical prototypes are sampled based on cluster centroids in a projected embedding space (see Figure \ref{fig:contracluster_overview}). The main idea is that cluster centroids in a low-dimensional space approximately represent the most discriminative samples. (for details see \ref{sec:stage2}).

In the third stage, we use the prototypes as noisy labeled data to perform \textit{prototype-based semi-supervised fine-tuning} (PB-SFT) that can increase the accuracy by leveraging both small noisy labeled data (i.e., prototypes) and large unlabeled data. PB-SFT can avoid the problem of over-fitting during fine-tuning with few labeled data. For leveraging unlabeled data, we adopt FixMatch \cite{sohn2020fixmatch}, one of the most successful single-stage semi-supervised learning method. (for details see \ref{sec:stage3}).

We empirically demonstrate that ContraCluster achieves new state-of-the-art results for standard benchmark datasets including CIFAR-10 \cite{krizhevsky2009learning}, STL-10 \cite{coates2011analysis}, and ImageNet-10 \cite{deng2009imagenet} (see Table \ref{tab:clustering_accuracy}). For CIFAR-10, ContraCluster achieves about 90.8\% accuracy that outperforms strong previous method such as DAC \cite{chang2017deep} (52.2\%), IIC \cite{ji2019invariant} (61.7\%), and SCAN \cite{van2020scan} (87.6\%) by a large margin. Note also that without labels, ContraCluster can achieve about 90.8\% accuracy that is comparable with the accuracy of supervised learning with full labels (95.8\%) \cite{pham2021autodropout}. Note that our method cannot be directly compared to deep clustering like \cite{huang2020deep, regatti2021consensus, li2021contrastive}, because ours does not repeat clustering procedure for new data, but inference classes with the trained model.

Our contributions are summarized as follows.
\begin{itemize}
  \item We propose novel unsupervised image classification method of robust prototype sampling.

  \item We empirically achieve new state-of-the-art results for standard benchmark datasets.
\end{itemize}

The rest of this paper is organized as follows. Section 2 discusses related work. In Section 3, we explain ContraCluster in detail. Section 4 presents our experimental results on the standard image benchmarks. The paper concludes in Section 5.

\section{Related Work}

\subsection{Self-supervised representation learning}
\paragraph{Task-specific self-supervised learning.}
Self-supervised learning aims to learn general representations from unlabeled data by performing a pretext task, then to reuse them in downstream tasks. For example, the pretext task includes context prediction \cite{doersch2015unsupervised}, jigsaw puzzle solving \cite{noroozi2016unsupervised}, image colorization \cite{zhang2016colorful}, and rotation prediction \cite{gidaris2018unsupervised}.

\paragraph{Contrastive self-supervised learning.}
Contrastive self-supervised learning extracts general representations from unlabeled data by using contrastive loss, which is based on the InfoMax principle \cite{tschannen2020mutual} that encourages the agreement between multiple views from an instance. It provides much higher quality visual representations in terms of the linear separability than the task-specific self-supervised learning. They include CPC \cite{oord2018representation}, DIM \cite{hjelm2019learning}, AM-DIM \cite{bachman2019learning}, MoCo \cite{he2020momentum}, and SimCLR \cite{chen2020simple}.

\subsection{Semi-supervised learning}
Semi-supervised learning enables to learn from small labeled data by leveraging large unlabeled data. Approaches to semi-supervised learning includes pseudo-labeling \cite{lee2013pseudo}, entropy minimization \cite{grandvalet2005semi}, and consistency regularization \cite{sajjadi2016regularization, tarvainen2017mean, laine2016temporal}. We mainly discuss consistency regularization in this paper.

\paragraph{Consistency regularization.}
Consistency regularization aims to use unlabeled data to regularize the cross-entropy loss with few labeled data. Its objective encourages a model to predict consistent class probabilities over stochastically transformed samples. It has been introduced in $\Pi$-Model \cite{sajjadi2016regularization} and further developed by MeanTeacher \cite{tarvainen2017mean, laine2016temporal}. Recently, advanced methods are introduced. They include MixMatch \cite{berthelot2019mixmatch}, UDA \cite{xie2020unsupervised}, ReMixMatch \cite{berthelot2020remixmatch}, FixMatch \cite{sohn2020fixmatch}.

\paragraph{Self-supervised pre-training-based.}
Recent work extends the self-supervised paradigm with more sophisticated fine-tuning techniques. For example, SimCLRv2 \cite{chen2020big} proposes to use the third stage of distillation after supervised fine-tuning with few labels. S4L \cite{zhai2019s4l} and SelfMatch \cite{kim2020selfmatch} show that semi-supervised fine-tuning further enhance the accuracy and label efficiency.

\subsection{Unsupervised classification / clustering}
\label{sec:unsupervised_classification}

\paragraph{From-scratch approach.}
Unsupervised image classification, or image clustering, can be broadly categorized into two: generative and discriminative. Generative approach attempts to learn general representations by using reconstruction or adversarial losses. This approach includes Autoencoder (AE)~\cite{bengio2007greedy}, GAN \cite{radford2015unsupervised}, VAE \cite{kingma2013auto}, and ClusterGAN \cite{ghasedi2019balanced}. In contrast, discriminative approach tries to learn general representations by using unsupervised loss that encourages proper cluster assignment in a label space. This approach includes DEC~\cite{xie2016unsupervised}, DAC \cite{chang2017deep}, DeepCluster \cite{caron2018deep}, IIC \cite{ji2019invariant}, DCCM \cite{wu2019deep}. However, these two approaches usually suffered from relatively low accuracy since they do not directly optimize representations in an embedding space.

\paragraph{Self-supervised pre-training-based.}
Most recently, the use of contrastive self-supervised pre-training has been proposed for unsupervised image classification or clustering. Since contrastive self-supervised pre-training aims to learn general representations by directly optimizing them in the embedding space, it has a huge potential to improve the clustering accuracy compared to the previous approaches. This approach includes SCAN \cite{van2020scan} and RUC \cite{park2020improving}. 

\paragraph{Comparison with other methods.}
SCAN \cite{van2020scan} proposes to fine-tune the SimCLR \cite{chen2020simple} pre-trained encoder by using an unsupervised loss that encourages the similarity of cluster assignment probabilities between a sample and \textit{$k$ nearest neighborhoods}. RUC \cite{park2020improving} proposes to further fine-tune the SCAN model by using \textit{interpolation consistency regularization} (e.g., MixMatch \cite{berthelot2019mixmatch}). In contrast to these methods, ContraCluster proposes to fine-tune the pre-trained encoder by using prototype-based semi-supervised fine-tuning (PB-SFT). To achieve this goal, we develop contrastive prototype sampling (CPS) that selects categorically-accurate prototypes to use as noisy labeled data for semi-supervised learning. Similar to ContraCluster, RUC selects clean labeled data by using confidence. However, ContraCluster selects clean labeled data represented by prototypes based on cluster centroids that can be discovered in embedding space.
    
\section{Method}

\begin{figure}[t]
    \begin{center}
    \includegraphics[width=0.45\textwidth]{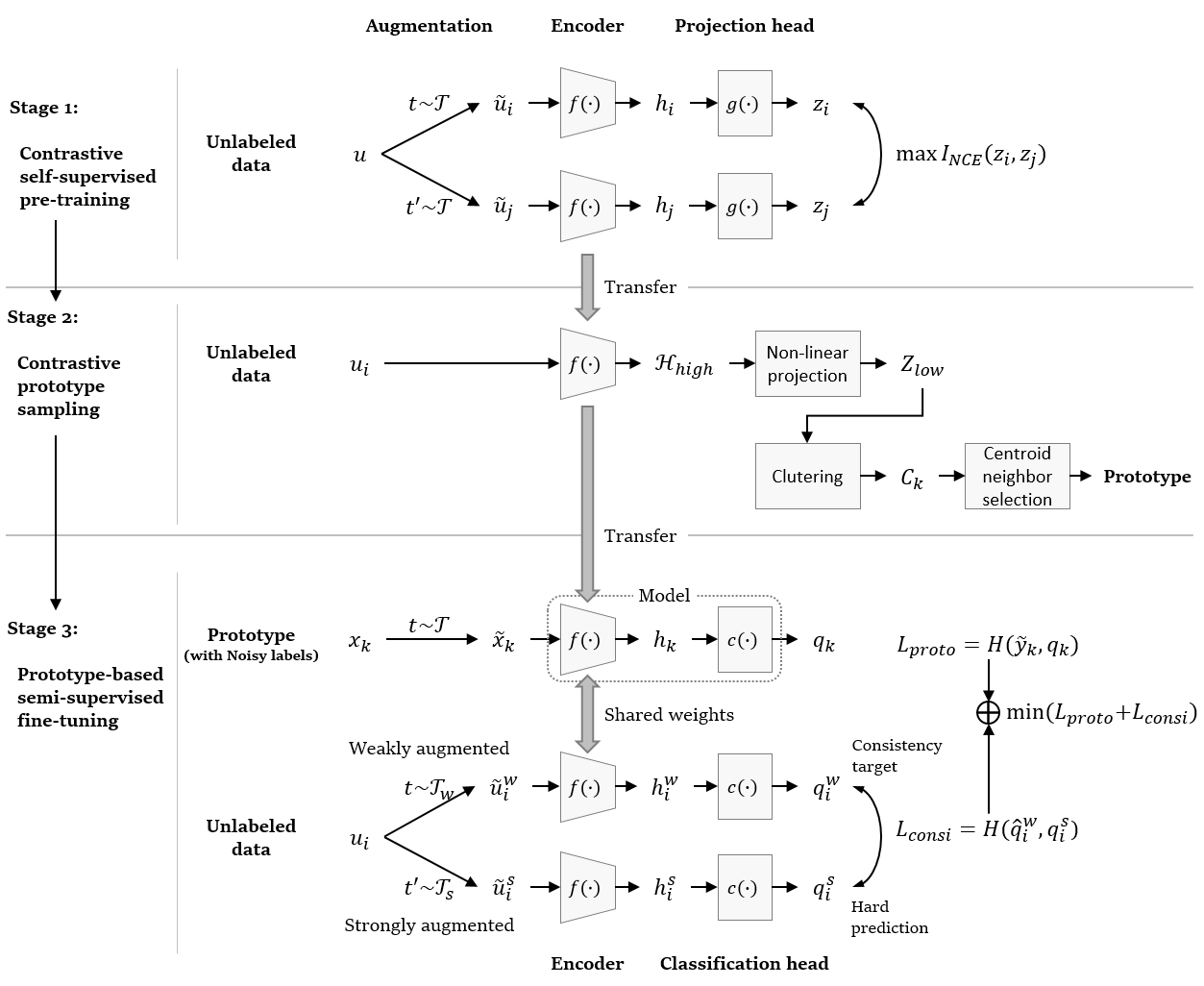}
    \end{center}
    \caption{The model architecture of ContraCluster.}
    \label{fig:contracluster_model_architecture}
\end{figure}

\begin{algorithm}[t]
	\caption{The ContraCluster 3-stage learning algorithm.}
	\label{algo:contracluster_algorithm}
	
	\begin{algorithmic}[1]
		\Require Unlabeled data $\mathcal{U}$
		\Require Randomly initialize encoder $f(\cdot)$, projection head $g(\cdot)$, and classification head $c(\cdot)$
		\State \# Stage 1: Contrastive self-supervised pre-training
		\For{$n=1$ to $E_{CPT}$}
		\For{$k=1$ to $B_{CPT}$}
		\State $u_k \sim \mathcal{U}$ \Comment{unlabeled batch}
		\State $t \sim \mathcal{T}, t' \sim \mathcal{T}$
		\State $\tilde{u}_{i} = t(u_k), \tilde{u}_{j} = t'(u_k)$ \Comment{transformation}
		\State $h_i = f(\tilde{u}_{i}), h_j = f(\tilde{u}_{j})$ \Comment{encoding}
		\State $z_i = g(h_i), z_j = g(h_j)$ \Comment{projection}
		\State $\mathcal{L}_{CPT} = -\log \frac{\textrm{exp}(\textrm{sim}(z_i,z_j)/\tau)}{\sum_{k=1}^{2B_{CPT}} \mathds{1}(k \neq i) \textrm{exp}(\textrm{sim}(z_i,z_k)/\tau)}$ \par \Comment{Eq. \ref{eq:contrastive_loss}}
		\EndFor
		\State $SGD(\eta_{CPT})$
		\EndFor
		\State \# Stage 2: Contrastive prototype sampling
		\For{$\forall u_i \in \mathcal{U}$}
		\State $\mathcal{H}_{high} \gets f(u_i)$
		\EndFor
		\State $\mathcal{Z}_{low} = projection(\mathcal{H}_{high}, N_{neigh}, N_{dim})$
		\State $\mathcal{C}_k = clustering(\mathcal{Z}_{low}, k_{part})$
		\State $\mathcal{P} = sampleCentroidNeighbors(\mathcal{C}_k, N_{proto})$
		\State \# Stage 3: Prototype-based semi-supervised fine-tuning
		\For{$n=1$ to $E_{SFT}$}
		\For {$k=1$ to $B_{SFT}$}
		\State $x_k \sim \mathcal{P}$ \Comment{prototype batch}
		\State $t \sim \mathcal{T}$
		\State $\tilde{x}_k = t(x_k)$
		\State $q_k = c(f(\tilde{x}_k))$
		\State $\mathcal{L}_{proto} = \frac{1}{B_{SFT}} \sum_{k=1}^{B_{SFT}} H(\hat{y}_k, q_k)$ \Comment{Eq. \ref{eq:prototype_loss}}
		\EndFor
		\For{$i=1$ to $\mu B_{SFT}$}
		\State $u_i \sim \mathcal{U}$ \Comment{unlabeled batch}
		\State $t \sim \mathcal{T}_{weak}, t' \sim \mathcal{T}_{strong}$
		\State $\tilde{u}_i^w = t(u_i)$ \Comment{weak augmentation}
		\State $\tilde{u}_i^s = t'(u_i)$ \Comment{strong augmentation}
		\State $q_i^w = c(f(\tilde{u}_i^w))$
		\State $q_i^s = c(f(\tilde{u}_i^s))$ \Comment{hard prediction}
		\State $\hat{q}_i^w = argmax(q_i^w)$ \Comment{consistency target}
		\State $m =  \mathds{1}(\textrm{max}(q_{i}^{w}) \geq c)$ \Comment{mask}
		\State $\mathcal{L}_{consi} = \frac{1}{\mu B_{SFT}} \sum_{i=1}^{\mu B_{SFT}} m H(\hat{q}_i^w, q_i^s)$ \par \Comment{Eq. \ref{eq:consistency_loss}}
		\EndFor
		\State $\mathcal{L}_{PB-SFT} = \mathcal{L}_{proto} + \lambda_{u} \mathcal{L}_{consi}$ \Comment{Eq. \ref{eq:sft_loss}}
		\State $SGD(\eta_{SFT})$
		\EndFor
		\Return encoder $f(\cdot)$, classification head $c(\cdot)$
	\end{algorithmic}
\end{algorithm}

In this section, we describe the details of ContraCluster. Figure \ref{fig:contracluster_model_architecture} shows the model architecture. It consists of three stages: (1) contrastive self-supervised pre-training (CPT), (2) contrastive prototype sampling (CPS), and (3) prototype-based semi-supervised fine-tuning (PB-SFT). The weight of the encoder pre-trained in the first stage is transferred in the following stages. In the final stage, the pre-trained encoder is further fine-tuned by using both small noisy labeled data (i.e., prototypes) and large unlabeled data. The model $p_{model}(y|x)$ consists of an encoder $f(\cdot)$ and a head $c(\cdot)$. The learning algorithm of ContraCluster is presented in Algorithm \ref{algo:contracluster_algorithm}.

\subsection{Contrastive self-supervised pre-training}
\label{sec:stage1}

ContraCluster aims to learn a linearly separable embedding space by using only unlabeled data in the first stage. For CPT, it adopts SimCLR \cite{chen2020simple}, one of the most effective methods. SimCLR learns representations by simultaneously encouraging two objectives: (1) maximizing the similarity between different views $\tilde{u}_i$ and $\tilde{u}_j$ from the same sample and (2) minimizing the similarity between different views $\tilde{u}_i$ and $\tilde{u}_k$ $(k \neq i)$ from different samples.

As shown in Figure \ref{fig:contracluster_model_architecture}, SimCLR consists of four components: (1) data augmentation $\mathcal{T}(\cdot)$, (2) encoder $f(\cdot)$, (3) projection head $g(\cdot)$, and (4) contrastive loss $\mathcal{L}_c$. The data augmentation uses random crop and color distortion. The encoder is ResNet-50 \cite{he2016deep}. The projection head is a two-layer MLP with dropout \cite{srivastava2014dropout} and ReLU activation. Finally, the contrastive loss is formulated as follows:
\begin{equation}
    \label{eq:contrastive_loss}
    \mathcal{L}_{CPT} = -\log \frac{\textrm{exp}(\textrm{sim}(z_i,z_j)/\tau)}{\sum_{k=1}^{2B_{CPT}} \mathds{1}(k \neq i) \textrm{exp}(\textrm{sim}(z_i,z_k)/\tau)}
\end{equation}
where $B_{CPT}$ is a batch size, $\mathds{1}(k \neq i)$ is an indicator function evaluating to 1 only if $k \neq i$, $\textrm{sim}(\cdot)$ is a similarity function, and $\tau$ is a temperature parameter scaling the similarity. The contrastive learning hyperparameters are listed in Table \ref{tab:hyperparameters}.

\subsection{Contrastive prototype sampling}
\label{sec:stage2}
CPS is to select highly accurate prototypical images from the embedding space of the first stage. For prototypes, CPS simply chooses $n$ nearest neighbors from the \textit{cluster centroids}. To do so, CPS first reduces the dimension of space ($\mathcal{H}_{high} \rightarrow \mathcal{Z}_{low}$) by using a non-linear dimensionality reduction algorithm such as UMAP \cite{mcinnes2018umap} and t-SNE \cite{van2008visualizing}. Then, k-means clustering \cite{kanungo2002efficient} ($\mathcal{C}_k$) applies on the projected embedding space (e.g., a 2-dimensional space) to find cluster centroids, assuming the groud-truth number of clusters is given, holding the same condition with previous works \cite{van2020scan, park2020improving} to compare. Finally, CPS selects $n$ nearest neighbors ($\mathcal{P}$) from the cluster centroids. We have empirically found out that non-linear dimensionality reduction is essential to increase the accuracy (see Table \ref{tab:ablation_study}). We conjecture that this is mainly due to the fact that classic clustering algorithms such as k-means suffer from the low accuracy problem because of the curse of dimensionality. Note that DBSCAN \cite{ester1996density}, an alternative to k-means, cannot be applied by lack of cluster centroids.

To determine the proper hyperparameter values for UMAP (e.g., \# of neighbors $N_{neigh}$, \# of dimension $N_{dim}$, etc.), we use Silhouette Coefficient \cite{rousseeuw1987silhouettes} (see Figure \ref{fig:silhouette_scores_prototype_accuracy}). The Silhouette Coefficient $(y-x)/max(x, y)$ is evaluated on the mean intra-cluster distance $(x)$ and the mean nearest-cluster distance $(y)$ for each sample. Since Silhouette Coefficient is calculated with clustering results (i.e., no ground-truths required), we can approximately choose the best hyperparameters of UMAP and k-means by using only unlabeled data. Hyperparameters are summarized in Table \ref{tab:hyperparameters}.

\subsection{Prototype-based semi-supervised fine-tuning}
\label{sec:stage3}

PB-SFT can further increase the accuracy by leveraging both small noisy labeled prototypes and (large) unlabeled data. We adopts FixMatch \cite{sohn2020fixmatch} that exploits augmentation-based consistency regularization for unlabeled data. It encourages the consistent prediction between weakly and strongly augmented examples. More specifically, its objective function is to minimize the cross entropy $H(p, q)$ between the prediction $\hat{q}_i^w$ of a weakly augmented input $\tilde{u}_i^w$ and the class probability distribution $q_i^s$ of a strongly augmented input $\tilde{u}_i^s$ (see Figure \ref{fig:contracluster_model_architecture}). The weak augmentation uses random crops and horizontal flips. The strong one adopts RandAugment (RA) \cite{cubuk2020randaugment}, an effective automated method. The classification head $c(\cdot)$ is a MLP of two layers with dropout \cite{srivastava2014dropout} and ReLU activation.

The loss function consists of prototype-based cross-entropy loss and consistency regularization loss: 
\begin{equation}
    \label{eq:sft_loss}
    \mathcal{L}_{PB-SFT} = \mathcal{L}_{proto} + \lambda_{u} \mathcal{L}_{consi}
\end{equation}

The prototype-based supervised loss $\mathcal{L}_{proto}$ is formulated as follows: 
\begin{equation}
    \label{eq:prototype_loss}
    \mathcal{L}_{proto} = \frac{1}{B_{SFT}} \sum_{k=1}^{B_{SFT}} H(\hat{y}_k, q_k)
\end{equation}
where $B_{SFT}$ is a fine-tuning batch size, $\hat{y}_i$ is a noisy label of a prototype.

The augmentation-based consistency regularization loss $\mathcal{L}_{consi}$ is formulated as follow: 
\begin{equation}
    \label{eq:consistency_loss}
    \mathcal{L}_{consi} = \frac{1}{\mu B_{SFT}} \sum_{i=1}^{\mu B_{SFT}} \mathds{1}(\textrm{max}(q_{i}^{w}) \geq c) H(\hat{q}_i^w, q_i^s)
\end{equation}
where  $c$ is a confidence threshold, $\mathds{1}(\textrm{max}(q_{i}^{w}) \geq c)$ is an indicator function, $B_{SFT}$ is a fine-tuning batch size, and $\mu$ is the ratio of prototypes and unlabeled samples in a batch.

For training, it utilizes Exponential Moving Average (EMA) \cite{tarvainen2017mean} with a weight decay for stable training and inference. The hyperparameters are described in Table \ref{tab:hyperparameters}. Many of them follows SimCLR \cite{chen2020simple} and FixMatch \cite{sohn2020fixmatch}, to make a fair comparison with the other methods.

\section{Experiments}

\subsection{Datasets}
We empirically validate ContraCluster using standard benchmark datasets: CIFAR-10 \cite{krizhevsky2009learning}, STL-10 \cite{coates2011analysis}, and ImageNet-10 \cite{deng2009imagenet}, as in Table \ref{tab:dataset_summary}.


\begin{table}[h]
    \begin{center}
    \begin{tabular}{m{2.0cm} m{1.3cm} m{1.7cm} m{1.5cm}}
    \toprule
    \noalign{\smallskip}
    Dataset & CIFAR10 & STL10 & ImageNet10 \\
    
    \cmidrule(lr){1-1} \cmidrule(lr){2-2} \cmidrule(lr){3-3} \cmidrule(lr){4-4}
    \noalign{\smallskip}
    Size & 32x32 & 96x96 & 224x224 \\
    Classes & 10 & 10 & 10 \\
    Train split & train & train+test & train \\
    Test split & test & train+test & train \\ 
    Train samples & 50,000 & 5,000+8,000 & 13,000 \\
    Test samples & 10,000 & 5,000+8,000 & 13,000 \\
    \bottomrule
    \end{tabular}
    \end{center}
    
    \caption{Summary of datasets.}
    \label{tab:dataset_summary}
\end{table}

\subsection{Hyperparameter setting}

\begin{table}[h]
    \begin{center}
    \begin{tabular}{m{3.2cm} m{1.3cm} m{1.3cm} m{1.3cm}}
    \toprule
    \noalign{\smallskip}
    Hyperparameters & CIFAR10 & STL10 & ImageNet10 \\
    
    \midrule
    \noalign{\smallskip}
    Temperature $\tau$ & 0.1 & 0.1 & 0.1 \\
    Batch size $B_{CPT}$ & 512 & 256 & 64 \\
    Optimizer & SGD & SGD & SGD \\
    Learning rate $\eta_{CPT}$ & 0.6 & 0.3 & 0.075 \\
    Max epoch $E_{CPT}$ & 1024 & 1024 & 1024 \\
    
    \cmidrule(lr){1-1} \cmidrule(lr){2-2} \cmidrule(lr){3-3} \cmidrule(lr){4-4}
    \noalign{\smallskip}
    \# of neigh. $N_{neigh}$ & 20 & 50 & 50 \\
    Projection dim. $N_{dim}$ & 2 & 2 & 2 \\
    Min. distance $D_{min}$ & 0.5 & 0.0 & 0.0 \\
    Similarity metric & correl. & correl. & correl. \\ 
    \# of proto. $N_{proto}$ & 250 & 1000 & 1000 \\
    
    \cmidrule(lr){1-1} \cmidrule(lr){2-2} \cmidrule(lr){3-3} \cmidrule(lr){4-4}
    \noalign{\smallskip}   
    Batch size $B_{SFT}$ & 64 & 64 & 64 \\
    Unlab. batch ratio $\mu$ & 7 & 7 & 7 \\
    Unlab. loss ratio $\lambda_l$ & 1 & 1 & 1 \\
    Confidence thre. $c$ & 0.95 & 0.95 & 0.95 \\
    Optimizer & SGD & SGD & SGD \\
    Learning rate $\eta_{SFT}$ & 0.03 & 0.03 & 0.03 \\
    Max epoch $E_{SFT}$ & 400 & 400 & 400 \\

    \bottomrule
    \end{tabular}
    \end{center}
    
    \caption{Hyperparameters of ContraCluster.}
    \label{tab:hyperparameters}
\end{table}

Table \ref{tab:hyperparameters} presents a complete list of the hyperparameters. Each partition of the table shows the values used in the stage one to three respectively. They are empirically determined.


\subsection{Hyperparameter selection for contrastive prototype sampling}

\begin{figure}[h]
    \begin{center}
    \includegraphics[width=1.0\linewidth, height=4.0cm]{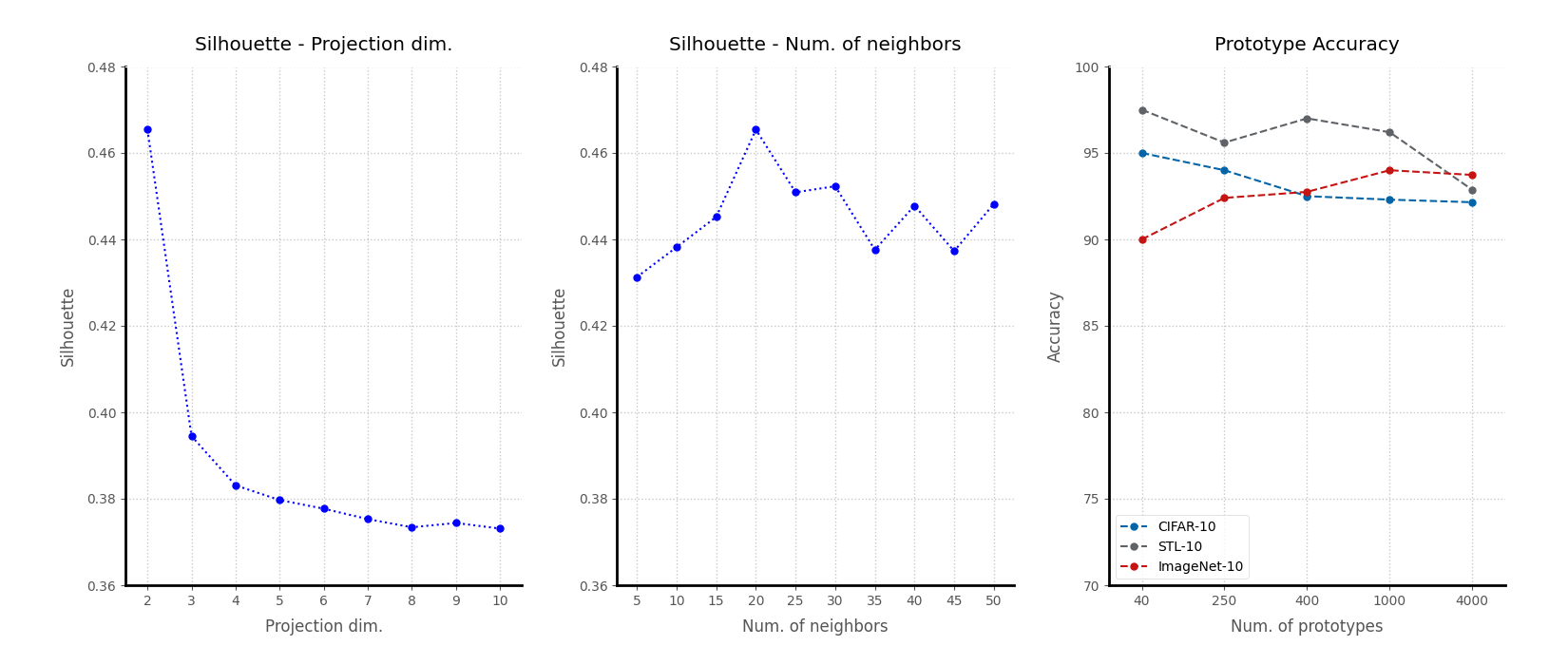}
    \end{center}
    \caption{Hyperparameter selection with Silhouette Coefficient for contrastive prototype sampling and prorotype accuracy of ContraCluster. (Left) w.r.t the projection dimension. (Center) w.r.t the number of neighbors. (Right) w.r.t prototype accuracy.}
    \label{fig:silhouette_scores_prototype_accuracy}
\end{figure}

To choose the proper hyperparamter values for CPS, we use Silhouette Coefficient. Figure \ref{fig:silhouette_scores_prototype_accuracy} shows the variation of it with respect to \# of neighbors $N_{neigh}$ and the projected dimension $N_{dim}$ of UMAP. For CIFAR-10, we choose 2 for $N_{dim}$ and 20 for $N_{neigh}$, where it is the highest (see Table \ref{tab:hyperparameters}).

\subsection{Unsupervised image classification accuracy}

\begin{table*}[t]
    \begin{center}
    \begin{tabular}{m{4.5cm} m{1.5cm} m{1.2cm} m{1.5cm} m{1.2cm} m{1.5cm} m{1.2cm}}
    \toprule
    & \multicolumn{2}{c}{\textbf{CIFAR-10}} & \multicolumn{2}{c}{\textbf{STL-10} (train+test)} & \multicolumn{2}{c}{\textbf{ImageNet-10}} \\
    \textbf{Method} & Acc.(\%) & NMI & Acc.(\%) & NMI & Acc.(\%) & NMI \\

    \cmidrule(lr){1-1} \cmidrule(lr){2-3} \cmidrule(lr){4-5} \cmidrule(lr){6-7}
    \noalign{\smallskip}
    \textbf{Supervised} (full labels) \cite{pham2021autodropout} & \textbf{95.8} & - & - & - & \textbf{91.4} & - \\

    \cmidrule(lr){1-1} \cmidrule(lr){2-3} \cmidrule(lr){4-5} \cmidrule(lr){6-7}
    \noalign{\smallskip}
    k-means* \cite{kanungo2002efficient} & 22.9 & 0.087 & 19.2 & 0.125 & - & - \\
    Spectral clustering* \cite{ng2002spectral} & 24.7 & 0.103 & 15.9 & 0.098 & - & - \\

    \cmidrule(lr){1-1} \cmidrule(lr){2-3} \cmidrule(lr){4-5} \cmidrule(lr){6-7}
    \noalign{\smallskip}
    Autoencoder (AE)* \cite{bengio2007greedy} & 31.4 & 0.234 & 30.3 & 0.250 & - & - \\
    DCGAN* \cite{radford2015unsupervised} & 31.5 & 0.265 & 29.8 & 0.210 & - & - \\
    ClusterGAN \cite{ghasedi2019balanced} & 41.2 & 0.323 & 42.3 & 0.335 & - & - \\

    \cmidrule(lr){1-1} \cmidrule(lr){2-3} \cmidrule(lr){4-5} \cmidrule(lr){6-7}
    \noalign{\smallskip}
    DEC* \cite{xie2016unsupervised} & 30.1 & 0.257 & 35.9 & 0.276 & - & - \\ 
    DAC* \cite{chang2017deep} & 52.2 & 0.400 & 47.0 & 0.366 & - & - \\
    DeepCluster* \cite{caron2018deep} & 37.4 & - & 33.4 & - & - & - \\
    DCCM \cite{wu2019deep} (only train set) & 62.3 & 0.496 & 48.2 & 0.376 & 71.0 & - \\
    IIC* \cite{ji2019invariant} & 61.7 & 0.511 & 59.6 & 0.496 & - & - \\

    \cmidrule(lr){1-1} \cmidrule(lr){2-3} \cmidrule(lr){4-5} \cmidrule(lr){6-7}
    \noalign{\smallskip}
    CC \cite{li2020contrastive} & 79.0 & 0.705 & 85.0 & 0.764 & 89.3 & 0.859 \\
    SCAN* \cite{van2020scan} (only train set) & 87.6$\pm$0.4 & 0.787 & 76.7$\pm$1.9 & 0.680 & - & - \\
    RUC (Conf.) \cite{park2020improving} & 90.3 & - & 86.7 & - & - & - \\

    \cmidrule(lr){1-1} \cmidrule(lr){2-3} \cmidrule(lr){4-5} \cmidrule(lr){6-7}
    \noalign{\smallskip}
    \textbf{ContraCluster} (avg.) & \textbf{90.8}$\pm$0.5 & \textbf{0.837} & \textbf{87.5}$\pm$0.3 & \textbf{0.784} & \textbf{90.2}$\pm$0.4 & \textbf{0.804} \\
    \textbf{ContraCluster} (max) & \textbf{91.7} & \textbf{0.857} & \textbf{87.9} & \textbf{0.787} & \textbf{90.5} & \textbf{0.809} \\

    \bottomrule
    \end{tabular}
    \end{center}
    \caption{Comparison of unsupervised image classification accuracy. }
    \label{tab:clustering_accuracy}
\end{table*}

Table \ref{tab:clustering_accuracy} shows a comparison of unsupervised image classification performance measured in accuracy (\%) and NMI (normalized mutual information) \cite{vinh2010information}. Asterisked ($^*$) results come from the SCAN \cite{van2020scan} paper, and the others from respective original publications. We provide both the mean and maximum accuracy of ContraCluster. The mean is computed by averaging five evaluations with different random seed numbers. It achieves state-of-the-art results for CIFAR-10, STL-10, and ImageNet-10.

\paragraph{CIFAR-10.} ContraCluster achieves a 90.8\% classification accuracy that outperforms DAC (52.2\%), IIC (61.7\%), and SCAN (87.6\%) by significant margins. Note that, without any labels, it is comparable with the accuracy of supervised learning with full labels (95.8\%). 

\paragraph{STL-10.} ContraCluster achieves a 87.5\% accuracy that also outperforms DAC (47.0\%), IIC (59.6\%), and SCAN (76.7\%) significantly.

\paragraph{ImageNet-10.} ContraCluster achieves a 90.5\% accuracy, which corresponds to outperform notable existing methods such as DCCM (71.0\%) and CC (89.3\%).

\subsection{Prototype accuracy}

%
%

Figure \ref{fig:silhouette_scores_prototype_accuracy} shows the variation of prototype accuracy with respect to the number of them. This figure shows that ContraCluster can select highly accurate prototypes (about 95.0\%) that can be used as noisy labeled data for PB-SFT. We choose 250 prototypes for CIFAR-10 (about 96.5\%), 1,000 for STL-10 (about 96.2\%), and 1,000 for ImageNet-10 (about 94.0\%), all based on empirical trials. For CIFAR-10 and STL-10, although 40 prototypes provide the highest accuracy (more than 95.0\%), sufficient number of prototypes (i.e., more than 100) is required for PB-SFT to achieve high clustering accuracy.

\subsection{Ablation study}

\begin{table}[h]
    \begin{center}
    \begin{tabular}{m{4.5cm} m{1.7cm}}
    \toprule
    \noalign{\smallskip}
    Method & Accuracy(\%) \\
    
    \cmidrule(lr){1-1} \cmidrule(lr){2-2}
    \noalign{\smallskip}
    ContraCluster w/o SimCLR & 29.0 (-61.8) \\
    ContraCluster w/o UMAP & 82.4 (-8.4) \\
    ContraCluster w/o FixMatch & 84.4 (-6.4) \\

    \cmidrule(lr){1-1} \cmidrule(lr){2-2}
    \noalign{\smallskip}
    ContraCluster & \textbf{90.8} \\
    
    \bottomrule
    \end{tabular}
    \end{center}
    
    \caption{Ablation study of ContraCluster for CIFAR-10.}
    \label{tab:ablation_study}
\end{table}

Table \ref{tab:ablation_study} shows an ablation study result of ContraCluster for CIFAR-10. It proves that the each stage is essential for achieving the stage-of-the-art results. ContaCluster w/o SimCLR means applying UMAP and k-means on raw pixels (i.e., a sample space) without performing CPT. Since it is very difficult to capture semantic information from high-dimensional raw pixels, it shows significant performance degradation. Without UMAP, ContraCluster does not provide the state-of-the-art accuracy because UMAP can effectively help find cluster centroids in a low-dimensional space. Note that it is one reason why trivial application of the projection head of SimCLR as prototype sampler is suboptimal. Finally, without FixMatch, we could not have achieved the state-of-the-art either. This shows that PB-SFT is effective to further increase the accuracy.

\subsection{Example results}

\begin{figure}[h]
    \begin{center}
    \includegraphics[width=1.0\linewidth]{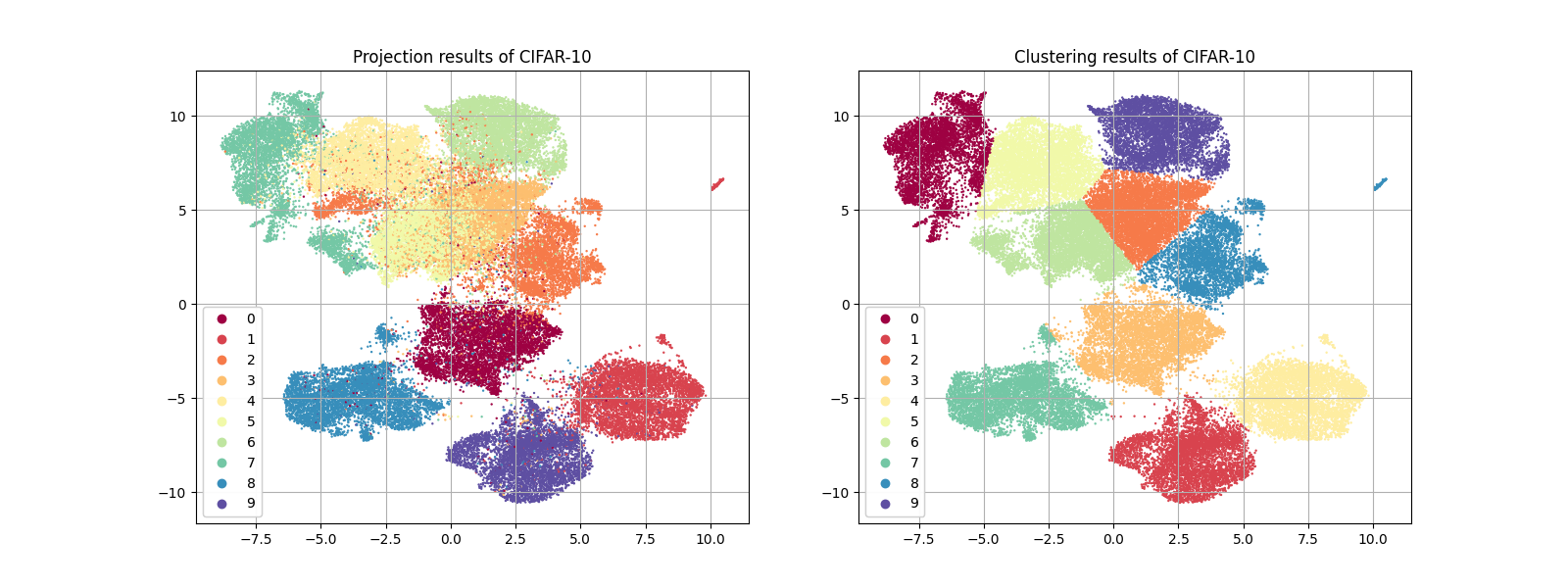}
    \end{center}

    \caption{Visualization of the embedding space. It is learned by ContraCluster for CIFAR-10. (Left) a projected embedding space by UMAP. (Right) a clustered embedding space by k-means.}
    \label{fig:embedding_space}
\end{figure}


\paragraph{Embedding space.}
Figure \ref{fig:embedding_space} shows an example of the embedding space learned by ContraCluster for CIFAR-10. The left side shows an projected embedding space of UMAP. The right side shows a clustered embedding space by k-means. We present the more examples in the appendix.


\paragraph{Clustering results.}
Figure \ref{fig:contracluster_main_result} shows an example of the final clustering result by ContraCluster. The example shows 90.8\% class-accurate clustering (see Table \ref{tab:clustering_accuracy}). We present more examples in the appendix.

%
%
%
%
%



\section{Conclusion}
We have presented ContraCluster, an unsupervised image classification method based on contrastive self-supervised learning. Combining the three stages, (1) contrastive self-supervised pre-training (CPT), (2) contrastive prototype sampling (CPS), and (3) prototype-based semi-supervised fine-tuning (PB-SFT), it build a high-performance classification pipeline without relying on labeled data. Our experimental evaluation indicates that it achieves new state-of-the-art results on CIFAR-10, STL-10, and ImageNet-10. 

\bibliographystyle{IEEEtran}
\bibliography{IEEEabrv,contracluster}
%
%
%

\end{document}